\documentclass[letterpaper, 10 pt, conference]{ieeeconf}  %

\IEEEoverridecommandlockouts                              %

\pdfoutput=1

\usepackage{amssymb}
\usepackage{amsmath}
\usepackage{amsfonts}       %
\usepackage{booktabs}       %
\usepackage[sort,compress]{cite}
\usepackage[utf8]{inputenc} %
\usepackage[T1]{fontenc}    %
\usepackage[pdftex, pdfstartview={FitV}, pdfpagelayout={TwoColumnLeft},bookmarksopen=true,plainpages = false, colorlinks=true, linkcolor=black, citecolor = black, urlcolor = black,filecolor=black , pagebackref=false,hypertexnames=false, plainpages=false, pdfpagelabels ]{hyperref}
\usepackage{balance}
\usepackage{url}            %
\usepackage{nicefrac}       %
\usepackage{microtype}      %
\usepackage[font=scriptsize]{caption}
\usepackage{subcaption}
\usepackage{graphicx}
\usepackage{epstopdf} %
\usepackage{mathtools}
\usepackage{marginnote}
\usepackage[dvipsnames]{xcolor}
\usepackage{float}
\usepackage[capitalize]{cleveref}
\usepackage[]{algorithmic}
\usepackage{makecell}
\usepackage{multirow}
\usepackage{paralist}
\floatstyle{boxed} \floatname{algorithm}{Algorithm}
\newfloat{algorithm}{t}{loa}[section]
\usepackage{xcolor}
\usepackage[nolist,nohyperlinks]{acronym}

\usepackage{amsmath}
\usepackage{color}
\usepackage{soul}
\usepackage{xspace}
\usepackage[capitalize]{cleveref}
\usepackage{dsfont}
\usepackage{bm}
\usepackage{bbm}
\usepackage{nicefrac}  %
\usepackage{amsfonts}
\usepackage{amssymb}
\usepackage{array}
\usepackage{mathtools}
\usepackage{caption}
\usepackage{subcaption}
\usepackage{fontawesome}
\usepackage{thmtools}
\usepackage{thm-restate}
\usepackage{lipsum}
\usepackage{pifont}
\usepackage{makecell}
\usepackage{multirow}
\usepackage{afterpage}
\usepackage{tabularx} 
\usepackage{url}
\usepackage{wrapfig}  %
\usepackage{textcomp}

\Crefname{figure}{Figure}{Figures}
\crefname{figure}{Figure}{Figures}
\crefname{table}{Table}{Tables}
\crefformat{table}{Table~#1}
\crefformat{equation}{Equation (#1)}
\Crefformat{equation}{Equation (#1)}
\crefformat{algorithm}{Algorithm~#1}
\crefformat{appendix}{Appendix~#1}
\Crefformat{appendix}{Appendix~#1}
\crefformat{appendices}{Appendices~#1}
\Crefformat{appendices}{Appendices~#1}

\newcommand\sdots{\hbox to 1em{.\hss.\hss.}} %
\DeclareMathAlphabet\mathbfcal{OMS}{cmsy}{b}{n}  %

\newif\ifcomments
\commentstrue

\ifcomments
	\newcommand{\dXX}[1]{\color{red}DK: (#1)\color{black}\xspace}  %
	\newcommand{\aXX}[1]{\color{OliveGreen}AT: (#1)\color{black}\xspace}  %
	\newcommand{\XX}[1]{\color{orange}JH: (#1)\color{black}\xspace}  %
\else
    \newcommand{\dXX}[1]{}  %
	\newcommand{\aXX}[1]{}  %
	\newcommand{\XX}[1]{}  %
\fi

\crefformat{equation}{(#2#1#3)}

\title{\LARGE \bf
Demonstration-Efficient Guided Policy Search via\\
Imitation of Robust Tube MPC
}

\author{Andrea Tagliabue, Dong-Ki Kim, Michael Everett, Jonathan P. How%
\thanks{All the authors are with the MIT Department of Aeronautics and Astronautics. \tt\{atagliab, dkkim93, mfe, jhow\}@mit.edu}%
}%

\begin{document}

\maketitle
\thispagestyle{empty}
\pagestyle{empty}

\begin{abstract}
We propose a demonstration-efficient strategy to compress a computationally expensive Model Predictive Controller (MPC) into a more computationally efficient representation based on a deep neural network and Imitation Learning (IL). By generating a Robust Tube variant (RTMPC) of the MPC and leveraging properties from the tube, we introduce a data augmentation method that enables high demonstration-efficiency, being capable to compensate the distribution shifts typically encountered in IL. Our approach opens the possibility of zero-shot transfer from a single demonstration collected in a nominal domain, such as a simulation or a robot in a lab/controlled environment, to a domain with bounded model errors/perturbations. Numerical and experimental evaluations performed on a trajectory tracking  MPC for a quadrotor show that our method outperforms strategies commonly employed in IL, such as DAgger and Domain Randomization, in terms of demonstration-efficiency and robustness to perturbations unseen during training.
\end{abstract}
\section{Introduction}

\ac{MPC}~\cite{borrelli2017predictive} enables impressive performance on complex, agile robots~\cite{lopez2019dynamic, lopez2019adaptive, li2004iterative, kamel2017linear, minniti2019whole, williams2016aggressive}. However, its computational cost often limits the opportunities for onboard, real-time deployment~\cite{lco2020170:online}, or takes away critical computational resources needed by other components governing the autonomous system. %
Recent works have mitigated \ac{MPC}'s computational requirements by relying on computationally efficient \acp{DNN}, which are leveraged to imitate task-relevant demonstrations generated by \ac{MPC}.
Such demonstrations are generally collected
via \ac{GPS}~\cite{levine2013guided, kahn2017plato, zhang2016learning, carius2020mpc} and \ac{IL}~\cite{kaufmann2020deep, ross2013learning, reske2021imitation} and are then used to train a \ac{DNN} via supervised learning. 

A common issue in existing IL methods (e.g., \ac{BC}~\cite{pomerleau1989alvinn, osa2018algorithmic, bojarski2016end}, \ac{DAgger}~\cite{ross2011reduction}) is that they require collecting a relatively large number of \ac{MPC} demonstrations, preventing training directly on a real robot and requiring a simulation environment that accurately represents the deployment domain.
One of the causes for such demonstration inefficiency is the need to take into account and correct for the compounding of errors in the learned policy~\cite{ross2011reduction}, which may otherwise create shifts (\textit{covariate shifts}) from the training distribution, with catastrophic consequences~\cite{pomerleau1989alvinn}. These errors can be caused by: 
\begin{inparaenum}[a)]
    \item approximation errors in the learned policy; 
    \item mismatches in the simulation dynamics due to modelling errors (i.e., sim2real gap); or
    \item model changes or disturbances that may not be present in a controlled training environment (e.g., simulation, or lab/factory when training on a real robot), but do appear during deployment in the real-world (i.e., lab2real gap).
\end{inparaenum} 
Approaches employed to compensate for such \textit{gaps}, such as \ac{DR}~\cite{peng2018sim, loquercio2019deep}, introduce further challenges, for example, requiring to apply disturbances/model changes during training.
\par

\begin{figure}
    \centering
    \includegraphics[width=\columnwidth]{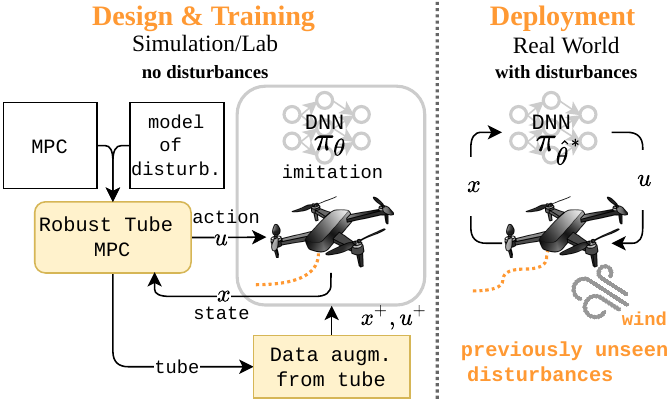}
    \caption{Overview of the proposed approach. To compress a computationally expensive MPC controller in a DNN-based policy $\pi_\theta$ in a demonstration efficient way, we generate a Robust Tube MPC controller using a model of the disturbances encountered in the deployment domain. We use properties of the tube to derive a data augmentation strategy which generates extra state-action pairs $(x^+, u^+)$, obtaining $\pi_{\hat{\theta}^*}$ via Imitation Learning. Our approach enables zero-shot transfer from a single demonstration collected in simulation (\textit{sim2real}) or a controlled environment (lab, factory, \textit{lab2real}).}
    \label{fig:sim_to_real_zero_shot_cover} \vspace*{-0.3in}
\end{figure}
In this work, we address the problem of generating a compressed MPC policy in a demonstration-efficient manner by providing a data augmentation strategy that systematically compensates for the effects of covariate shifts that might be encountered during real-world deployment. Specifically, our approach relies on a model of the perturbations/uncertainties encountered in a deployment domain, which is used to generate a Robust Tube version of the given MPC (RTMPC) to guide the collection of demonstrations and the data augmentation. An overview of the strategy is provided in \cref{fig:sim_to_real_zero_shot_cover}.
Our approach is related to the recent LAG-ROS framework~\cite{tsukamoto2021learning}, which provides a learning-based method to compress a planner in a \ac{DNN} by extracting relevant information from the robust tube. 
LAG-ROS emphasizes the importance of nonlinear contraction-based controllers (e.g., CV-STEM~\cite{tsukamoto2020neural}) to obtain robustness and stability guarantees. In a complementary way, our work emphasizes minimal requirements - namely a tube and a data augmentation strategy - to achieve demonstration efficiency and robustness to real-world conditions. By decoupling these aspects from the need for complex nonlinear models and control strategies, we greatly simplify the controller design and reduce the computational complexity, which enables  \textit{lab2real} transfers.
\par
\textbf{Contributions.} Via numerical comparison with previous \ac{IL} methods and experimental validations, we show that demonstration efficiency can be achieved in MPC-compression by generating a corresponding \ac{RTMPC} and using the tube to guide the data augmentation strategy for \ac{IL} methods (DAgger, BC). %
To this end, we propose a data-sparse, computationally efficient (i.e., scales linearly in state size) adversarial \ac{SA} strategy for data augmentation. We highlight that the proposed approach, for example, can be used to train the robot in a low-fidelity simulation environment %
while achieving robustness to real-world perturbations unseen during the training phase.  %
We validate the proposed approach by providing the first experimental (hardware) demonstration of zero-shot transfer of a DNN-based trajectory tracking controller for an aerial robot, learned from a \textit{single} demonstration, in an environment (low-fidelity simulation) without disturbances, and transferred to an environment with wind-like disturbances.

\section{Related Work}
\textbf{MPC-like policy compression for mobile robots via IL and GPS.}
\ac{IL} methods have found application in multiple robotics tasks. The works in~\cite{ross2013learning,pan2017agile} use \ac{DAgger}~\cite{ross2011reduction} to control aerial and ground robots, while~\cite{kaufmann2020deep} uses a combination of \ac{DAgger} and \ac{DR} to learn to perform acrobatic maneuvers with a quadrotor. %
Similarly, \ac{GPS} methods have been demonstrated in simulation for navigation  of a multirotor~\cite{zhang2016learning, kahn2017plato}, and to control a legged robot~\cite{carius2020mpc, reske2021imitation}. These methods achieve impressive performance, but at the cost of requiring multiple demonstrations to execute a single trajectory, and do not explicitly take into account the effects of disturbances encountered in the deployment domain.

\textbf{Sample efficiency and robustness in IL.}
Robustness in \ac{IL}, required to deal with distribution shifts caused by the sim2real (model errors) or lab2real transfer (model changes, external disturbances), has been achieved
\begin{inparaenum}[a)]
\item by modifying the training domain so that its dynamics match the deployment domain, as done in \ac{DR}~\cite{peng2018sim, loquercio2019deep, farchy2013humanoid, chebotar2019closing}, or
\item by modifying the actions of the expert, so that the state distribution during training matches the one encountered at deployment, as proposed in \cite{laskey2017dart, laskey2018and, hanna2017grounded, desai2020imitation}.
\end{inparaenum}
Although effective, these approaches do not leverage extra information available in the \ac{RTMPC}, thus requiring a larger number of demonstrations. 

\textbf{Data augmentation in IL.}
Data augmentation is a commonly employed robustification strategy in \ac{IL}. Most approaches focus on reducing overfitting in the high-dimensional policy input space (e.g., images), by applying noise~\cite{florence2019self}, transformations~\cite{hendrycks2019augmix} or adversarially-generated perturbations~\cite{shuadversarial, antotsiou2021adversarial}, while maintaining the corresponding action label unchanged. Data augmentation is also employed to reduce covariate shift in self-driving %
by generating transformed observations~\cite{toromanoff2018end, amini2020learning, bojarski2016end} with the corresponding action label computed via a feedback controller. These approaches do not directly apply to our context, as they do not rely on \ac{RTMPC} and we assume available state estimate. %
Aligned to our findings,~\cite{levine2013guided, carius2020mpc} observe that adding extra samples from the tube of an existing Iterative-\ac{LQR} can achieve increased demonstration efficiency in \ac{GPS}. 
Compared to these, thanks to \ac{RTMPC}, we can additionally consider the effects of disturbances encountered in the sim2real or lab2real transfer, providing additional robustness.

\section{Method}
This section explains the given MPC expert and its Robust-Tube variant, which we leverage to design a data augmentation strategy. We additionally cast the demonstration-efficiency challenge in \ac{IL} as a robust \ac{IL} problem in the context of transferring a policy between two different domains, and we present the \ac{SA} strategy to improve demonstration efficiency and robustness of MPC-guided policies learned via \ac{IL}.  

\subsection{MPC and Robust Tube MPC demonstrator}
\label{sec:rtmpc}
\textbf{Model predictive trajectory tracking controller.} We assume a trajectory tracking linear \ac{MPC}~\cite{borrelli2017predictive} is given that controls a system subject to bounded uncertainty. The linearized, discrete-time model of the system is:
\begin{equation}
    \check x_{t+1} = A \check x_{t} + B \check u_{t} + w_{t}
\label{eq:linearized_dynamics}
\end{equation}
where $x \in \mathbb{X} \subseteq \mathbb{R}^{n_x}$ represents the state (size $n_x$), $u \in \mathbb{U} \subseteq \mathbb{R}^{n_u}$ are the commanded actions (size $n_u$), and $w \in \mathbb{W} \subseteq \mathbb{R}^{n_x}$ is an additive perturbation/uncertainty. $A \in \mathbb{R}^{n_x \times n_x}$ and $B \in \mathbb{R}^{n_x \times n_u}$ represent the nominal dynamics; $\check \cdot$ denotes internal variables of the optimization.  
At every discrete timestep $t$ is given an estimate of the state of the actual system $x_t$ and a reference trajectory $x^{\text{ref}}_{0,t}, \dots, x^{\text{ref}}_{N,t}$. The controller computes a sequence of $N$ actions  $\check u_{0}(x_t), \dots, \check u_{N-1}(x_t)$, subject to state and input constraints $\mathbb{X}, \mathbb{U}$, and executes the first computed optimal action $u_t = {\check u^{\text{MPC,*}}_{0}}(x_t)$ from the sequence. The optimal sequence of actions is computed by minimizing the value function (dropping $x_t$ in the notation):
\begin{equation}
    V_t(\check u_{0},..., \check u_{N-1}, \check x_0\!)\!=\!\!\sum_{n=0}^{N-1}\!(e_n^T Q e_n +\check u_n^T R \check u_n)+ e_N^T P e_N\!,
\end{equation} 
where $e_{n} = (\check x_n - x^{\text{ref}}_{n,t})$. Matrices $Q$ (size $n_x \times n_x$) and $R$ (size $n_u \times n_u$) are user-selected, positive definite weights that define the stage cost, while $P$ (size $n_x \times n_x$, positive definite) 
represents the terminal cost. The prediction horizon is an integer $N \geq 1 $. The system is additionally subject to constraints $\check x_0 =x_t$; the predicted states $\check x_{1}, \dots, \check x_{N}$ are obtained from the model in \cref{eq:linearized_dynamics} assuming the disturbance $w=0$. %
The optimization problem is solved again at every timestep, executing the newly recomputed optimal action.

\begin{figure}
    \centering
    \includegraphics[width=0.6\columnwidth]{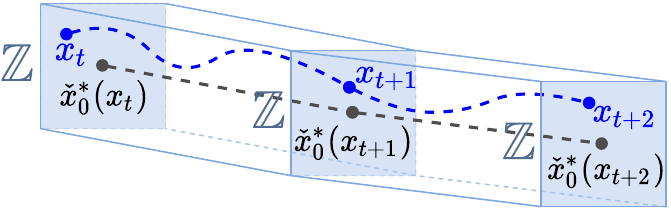}
    \caption{Illustration of the robust control invariant tube $\mathbb{Z}$ centered around the optimal reference $\check x_0^*(x_t)$ computed by RTMPC for each state of the system $x_t$.}
    \label{fig:tube_illustration} \vspace*{-0.25in}
\end{figure}

\textbf{Robust Tube MPC.} Given the \ac{MPC} expert, we generate a Robust Tube variant using~\cite{mayne2005robust}. At every discrete timestep $t$, the \ac{RTMPC} operates in a similar way as MPC, but it additionally accounts for the effects of $w$ by introducing a feedback policy, called  \textit{ancillary controller}
\begin{equation}
 u_t =  \check u_0^*(x_t) +  K(x_t - \check x_0^*(x_t)),
 \label{eq:ancillary_controller}
\end{equation}
where $u_t$ represents the executed action. The quantities $\check u_0^*(x_t)$ represents an optimal, feedforward action, and $\check x_0^*(x_t)$ is an optimal reference, and are computed by the \ac{RTMPC} given the current state estimate $x_t$. The ancillary controller ensures that the state $x$ of the controlled system remains inside a set (\textit{tube}) $\mathbb{Z} \in \mathbb{R}^{n_x}$, centered around $\check x_0^*(x_t)$, for every possible realization of $w \in \mathbb{W}$. The quantities $\check u_0^*(x_t)$ and $\check x_0^*(x_t)$ are obtained by solving (dropping the dependence on $x_t$)
\begin{equation}
   \check x_0^*, \check u_{0}^*, \dots = {\arg \min}_{\check x_0, \check u_{0},\dots, \check u_{N-1}}V_t( \check u_{0},\dots, \check u_{N-1}, \check x_0)
\end{equation}
under the constraint that $x_t \in \check x_0 + \mathbb{Z}$. As in the original MPC formulation, the optimization problem is additionally subject to the given input and actuation constraints, tightened by an amount that takes into account the effects of the disturbances. %
The gain matrix $K \in \mathbb{R}^{n_u \times n_x}$ in \cref{eq:ancillary_controller} is computed such that $A_K := A + B K$ is stable, for example by solving the infinite-horizon, discrete-time \ac{LQR} problem using ($A$, $B$, $Q$, $R$). 
The set $\mathbb{Z}$ has constant size, and  it determines the shape/width of the tube. It is defined as disturbance invariant set for the closed-loop system $A_K$, and satisfies the property that $\forall x_t \in \mathbb{Z}$, $\forall w_t \in \mathbb{W}$, $\forall t \in \mathbb{N}^+$, $x_{t+1} = A_K x_t + w_t \in \mathbb{Z}$. In practice, $\mathbb{Z}$ can be computed offline using $A_K$ and the model of the disturbance $\mathbb{W}$ via ad-hoc algorithms~\cite{borrelli2017predictive, mayne2005robust}, or can be learned from data~\cite{fan2020deep}.
The set $\mathbb{Z}$ and the ancillary controller in \cref{eq:ancillary_controller} ensure (see~\cite{mayne2005robust}) that, given a state $x_t \in \check x_0^*(x_t)+\mathbb{Z}$, the perturbed system in \cref{eq:linearized_dynamics} will remain in the tube centered around the trajectory of $\check x_0^*$, no matter the disturbance realization $w \in \mathbb{W}$, as shown in \cref{fig:tube_illustration}. This additionally implies that the tube represents a model of the states  that the system may visit when subject to the disturbances in $\mathbb{W}$. The ancillary controller provides a computationally-effective way to generate a control action to counteract such perturbations.

\subsection{Covariate shift in sim2real and lab2real transfer}
This part describes the demonstration-efficiency issue in \ac{IL} as the ability to efficiently predict and compensate for the effects of covariate shifts during real-world deployment. We assume that the causes of such distribution shifts can be modeled as additive state perturbations/uncertainties $w \in \mathbb{W}$ encountered in the deployment domains. %

\label{sec:transfer}

\textbf{Policies and state densities.} 
We model the dynamics of the real system as Markovian and stochastic~\cite{sutton2018reinforcement}.
The stochasticity with respect to state transitions is introduced by unknown perturbations, assumed to be additive (as in \cref{eq:linearized_dynamics}) and belonging to the bounded set $\mathbb{W}$, sampled under a (possibly unknown) probability distribution. These perturbations capture the effects of noise, approximation errors in the learned policy, model changes and other disturbances acting on the system. 
Two different domains $\mathcal{D}$ are considered: a training domain $\mathcal{S}$ (\textit{source}) and a deployment domain $\mathcal{T}$ (\textit{target}). The two domains differ in their transition probabilities, effectively representing the sim2real or lab2real settings. 
We additionally assume that the considered system is controlled by a deterministic policy $\pi_\theta : \mathbb{X} \times \mathbb{X}^\text{ref}_{0, \dots, N} \rightarrow \mathbb{U}$, where $\mathbb{X}^\text{ref}_{0, \dots, N}$ represents the reference trajectory. 
Given $(x^{\text{ref}}_{0,t}, \dots, x^{\text{ref}}_{N,t}) \in \mathbb{X}^\text{ref}_{0, \dots, N}$, the resulting transition probability is $p(x_{t+1}|x_t, u_t = \pi_\theta(x_t, x^{\text{ref}}_{0,t}, \dots, x^{\text{ref}}_{N,t}), \mathcal{D})$, denoted $p_{\pi_\theta, \mathcal{D}}(x_{t+1}|x_t)$ to simplify the notation.
The probability of collecting a $T$-step trajectory $\xi = (x_0, u_0, x_1, u_1, \dots, x_{T-1})$ given a generic policy $\pi_\theta$ in $\mathcal{D}$ is 
    $p(\xi|\pi_\theta, \mathcal{D}) = p(x_0) \prod_{t=0}^{T-1} p_{\pi_\theta, \mathcal{D}}(x_{t+1}|x_t)$,
where $p(x_0)$ represents the initial state distribution. 

\textbf{Robust IL objective.} Following~\cite{laskey2017dart}, given an expert \ac{RTMPC} policy $\pi_{\theta^*}$, the objective of \ac{IL} is to find parameters $\theta$ of $\pi_\theta$ that minimize a distance metric $J(\theta, \theta^*|\xi)$. This metric captures the differences between the actions generated by the expert $\pi_{\theta^*}$ and the action produced by the learner $\pi_\theta$ across the distribution of trajectories induced by the learned policy $\pi_\theta$, in the perturbed target domain $\mathcal{T}$:
\begin{equation}
    \hat{\theta}^* = \text{arg}\min_{\theta} \mathbb{E}_{p(\xi|\pi_\theta, \mathcal{T})}J(\theta, \theta^*|\xi)
    \label{eq:il_obj_target}
\end{equation}
A choice of distance metric that we consider in this paper is the MSE loss: $J(\theta, \theta^*|\xi) = \frac{1}{T}\sum_{t=0}^{T-1}\| \pi_\theta(x_t) - \pi_{\theta^*}(x_t)\|_2^2$.

\textbf{Covariate shift due to sim2real and lab2real transfer.}
Since in practice we do not have access to the target environment, our goal is to try to solve~\cref{eq:il_obj_target} by finding an approximation of the optimal policy parameters $\hat{\theta}^*$ in the source environment: 
\begin{equation}
    \hat{\theta}^* = \text{arg}\min_{\theta} \mathbb{E}_{p(\xi|\pi_\theta, \mathcal{S})}J(\theta, \theta^*|\xi).
    \label{eq:il_obj_source}
\end{equation}
The way this minimization is solved depends on the chosen \ac{IL} algorithm. The performance of the learned policy in target and source domain can be related via: 
\begin{equation}
\begin{multlined}
\hspace*{-0.45in}    \mathbb{E}_{p(\xi|\pi_\theta, \mathcal{T})}J(\theta, \theta^*|\xi) = \\
    \underbrace{
    \mathbb{E}_{p(\xi|\pi_\theta, \mathcal{T})}J(\theta, \theta^*|\xi) - 
    \mathbb{E}_{p(\xi|\pi_\theta, \mathcal{S})}J(\theta, \theta^*|\xi)}_{\text{covariate shift due to transfer}} \\ + 
    \underbrace{\mathbb{E}_{p(\xi|\pi_\theta, \mathcal{S})}J(\theta, \theta^*|\xi)}_{\text{\ac{IL} objective}},
\end{multlined}
\end{equation}
which clearly shows the presence of a covariate shift induced by the transfer. The last term corresponds to the objective minimized by performing \ac{IL} in $\mathcal{S}$. Attempting to solve \cref{eq:il_obj_target} by directly optimizing \cref{eq:il_obj_source} (e.g., via \ac{BC}~\cite{pomerleau1989alvinn}) provides no guarantees of finding a policy with good performance in $\mathcal{T}$. 

\textbf{Compensating transfer covariate shift via Domain Randomization.}
A well known strategy to compensate for the effects of covariate shifts between source and target domain is Domain Randomization (\ac{DR}) ~\cite{peng2018sim}, which modifies the transition probabilities of the source $\mathcal{S}$ by trying to ensure that the trajectory distribution in the modified training domain $\mathcal{S}_\text{DR}$ matches the one encountered in the target domain: $p(\xi|\pi_\theta, \mathcal{S}_\text{DR}) \approx p(\xi|\pi_\theta, \mathcal{T})$.
This is in practice done by sampling perturbations $w \in \mathbb{W}$ according to some knowledge/hypotheses on their distribution $p_\mathcal{T}(w)$ in the target domain~\cite{peng2018sim}, obtaining the perturbed trajectory distribution $p(\xi|\pi_\theta, \mathcal{S}, w)$. The minimization of \cref{eq:il_obj_target} can then be approximately performed by minimizing instead:
\begin{equation}
\label{eq:il_dr_modified_source}
    \mathbb{E}_{p_\mathcal{T}(w)}[\mathbb{E}_{p(\xi|\pi_\theta, \mathcal{S}, w)}J(\theta, \theta^*|\xi)].
\end{equation}
This approach, however, requires the ability to apply disturbances/model changes to the system, which may be unpractical e.g., in the \textit{lab2real} setting, and may require a large number of demonstrations due to the need to sample enough state perturbations $w$.

\subsection{Covariate shift compensation via Sampling Augmentation}
\label{sec:sa}
We propose to mitigate the covariate shift introduced by the compression procedure 
not only by collecting demonstrations from the \ac{RTMPC}, but by using additional information computed in the controller.
Unlike \ac{DR}, the proposed approach does not require to explicitly apply disturbances in the training phase. During the collection of a trajectory $\xi$ in the source domain $\mathcal{S}$, we utilize instead the tube computed by the \ac{RTMPC} demonstrator to obtain knowledge of the states that the system may visit when subjected to perturbations. Given this information, we propose a state sampling strategy, called \acf{SA}, to extract relevant states from the tube. The corresponding actions are provided at low computational cost by the demonstrator. The collected state-actions pairs are then included in the set of demonstrations used to learn a policy via \ac{IL}. The following paragraphs frame the tube sampling problem in the context of covariate shift reduction in \ac{IL}, and present two tube sampling strategies. %

\textbf{RTMPC tube as model of state distribution under perturbations.}
The key intuition of the proposed approach is the following. We observe that, although the density $p(\xi|\pi_\theta,\mathcal{T})$ is unknown, an approximation of its support $\mathfrak{R}$, given a demonstration $\xi$ collected in the source domain $\mathcal{S}$, is known. Such support corresponds to the tube computed by the \ac{RTMPC} when collecting $\xi$:
\begin{equation}
    \mathfrak{R}_{\xi^+|\pi_{\theta^*},\xi} = \bigcup_{t=0}^{T-1}\{\check x_0^*(x_t) + \mathbb{Z}\}.
    \label{eq:tube}
\end{equation}
where $\xi^+$ is a trajectory in the tube of $\xi$.
This is true thanks to the ancillary controller in \cref{eq:ancillary_controller}, which guarantees that the system remains inside \cref{eq:tube} for every possible realization of $w \in \mathbb{W}$. The ancillary controller additionally provides a way to easily compute the actions to apply for every state inside the tube. Let $x^+$ be a state inside the tube computed when the system is at $x_t$ (formally $x^+ \in \check x_0^*(x_t) + \mathbb{Z}$), then the corresponding robust control action $u^+$ is simply:
\begin{equation}
u^+ = \check u_0^*(x_t) +  K(x^+ - \check x_0^*(x_t)).
\label{eq:tubempc_feedback_policy}
\end{equation}
For every timestep in $\xi$, extra state-action samples $(x^+, u^+)$ collected from within the tube can be used to augment the dataset employed for empirical risk minimization, obtaining a way to approximate the expected risk in the domain $\mathcal{T}$ by only having access to demonstrations collected in $\mathcal{S}$: 
\begin{equation}
\begin{multlined}
    \mathbb{E}_{p(\xi|\pi_\theta, \mathcal{T})}J(\theta, \theta^*|\xi) \approx \\
    \mathbb{E}_{p(\xi|\pi_\theta, \mathcal{S})}[J(\theta, \theta^*|\xi) +  \mathbb{E}_{p(\xi^+|\pi_{\theta^*},\xi)}J(\theta, \theta^*|\xi^+)].
    \label{eq:sampling_augmentation}
\end{multlined}
\end{equation}
The demonstrations in the source domain $\mathcal{S}$ can be collected using existing techniques, such as \ac{BC} and \ac{DAgger}.

\textbf{Tube approximation and sampling strategies.}
\begin{figure}
    \centering
    \includegraphics[width=0.8\columnwidth]{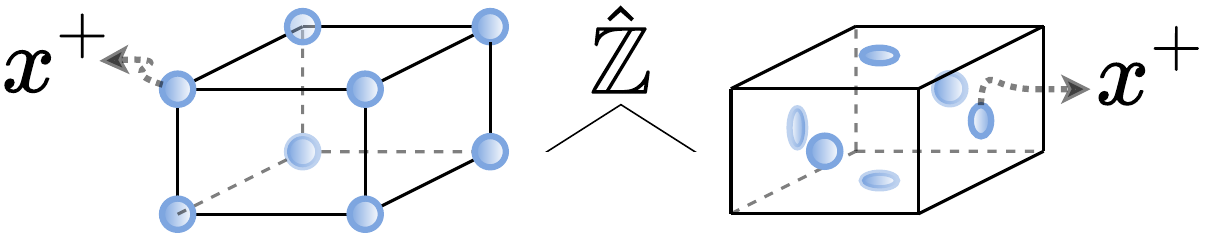}
    \caption{The two alternative strategies evaluated to sample extra state/actions pairs from an approximation of the tube of the \ac{RTMPC} expert: dense (left)  and sparse (right).}
    \label{fig:tube_sampling_strategies} \vspace*{-0.25in}
\end{figure}
In practice, the set $\mathbb{Z}$ may have arbitrary shape (not necessarily politopic), and the density $p(\xi^+|\xi, \pi_{\theta^*})$ may not be available, making difficult to establish where/which states to sample in order to derive a data augmentation strategy. We proceed by approximating $\mathbb{Z}$ as an hyper-rectangle $\hat{\mathbb{Z}}$, outer approximation of the tube. We consider an adversarial approach to the problem by sampling from the states visited under worst-case perturbations. We investigate two strategies, shown in~\cref{fig:tube_sampling_strategies}, to obtains state samples $x^+$ at every state $x_t$ in $\xi$:
\begin{inparaenum}[i)]
\item \textbf{dense sampling}: sample extra states from the vertices of $\check x_0^*(x_t) + \hat{\mathbb{Z}}$. The approach produces $2^{n_x}$ extra state-action samples. It is more conservative, as it produces more samples, but more computationally expensive.
\item \textbf{sparse sampling}: sample one extra state from the center of each \textit{facet} of $\check x_0^*(x_t) + \hat{\mathbb{Z}}$, producing  $2n_x$ additional state-action pairs. It is less conservative and more computationally efficient.
\end{inparaenum}

\section{Results}
\subsection{Evaluation approach}
\textbf{MPC for trajectory tracking on a multirotor.} We evaluate the proposed approach in the context of trajectory tracking control for a multirotor, using the controller proposed in \cite{kamel2017linear}, modified to obtain a \ac{RTMPC}. We model $\mathbb{W}$ under the assumption that the system is subject to force-like perturbations up to $30\%$ of the weight of the robot (approximately the safe physical limit of the robot). The tube $\mathbb{Z}$ is approximated via Monte-Carlo sampling of the disturbances in $\mathbb{W}$, evaluating the state deviations of the closed loop system $A_K$. The derived controller generates tilt (roll, pitch) and thrust commands ($n_u = 3$) given the state of the robot ($n_x=8$ consisting of position, velocity and tilt) and the reference trajectory. The reference is a sequence of desired positions and velocities for the next $3$s, discretized with sampling time of $0.1$s (corresponding to a planning horizon of $N=30$, and $180$-dim. vector). The controller takes into account position constraints (e.g., available 3D flight space), actuation limits, and velocity/tilt limits. %

\textbf{Policy architecture.} The compressed policy is a $2$-hidden layers, fully connected DNN, with $32$ neurons per layer, and ReLU activation function. The total input dimension of the DNN is $188$ (position, velocity, current tilt expressed in an inertial frame, and the desired reference trajectory). The output dimension is $3$ (desired thrust and tilt expressed in an inertial frame). We rotate the tilt output of the DNN in body frame to avoid taking into account yaw, which is not part of the optimization problem \cite{kamel2017linear}, not causing any relevant computational cost. We additional apply the non-linear attitude compensation scheme as in \cite{kamel2017linear}.

\textbf{Training environment and training details.} Training is performed in a custom-built non-linear quadrotor simulation environment, where the robot follows desired trajectories, starting from randomly generated initial states, centered around the origin. Demonstrations are collected with a sampling time of $0.1$s and training is performed for $50$ epochs via the ADAM \cite{kingma2014adam} optimizer, with a learning rate of $0.001$.

\textbf{Evaluation details and metrics.} We apply the proposed \ac{SA} strategies to \ac{DAgger} and \ac{BC}, and compare their performance against the two without \ac{SA}, and the two combined with \ac{DR}. Target and source domain differs due to perturbations sampled from $\mathbb{W}$ in target. %
During training with \ac{DR} we sample disturbances from the entire $\mathbb{W}$. 
In all the comparisons, we set the probability of using actions of the expert $\beta$, hyperparameter of DAgger \cite{ross2011reduction}, to be $1$ at the first demonstration, and $0$ otherwise (as this was found to be the best performing setup). We monitor:
\begin{inparaenum}[i)]
    \item \textit{robustness (Success Rate)}, as the percentage of episodes where the robot never violates any state constraint;
    \item \textit{performance (MPC Stage Cost)}, as $\sum_{t} e_t^T Q e_t + u_t^T R u_t$ along the trajectory. 
\end{inparaenum}

\subsection{Numerical evaluation of demonstration-efficiency, robustness and performance for tracking a single trajectory}

\begin{figure}
\captionsetup[sub]{font=footnotesize}
\centering
\begin{subfigure}{\columnwidth}
    \centering
    \includegraphics[width=\columnwidth, trim={2cm 0 0cm 0.5cm},clip]{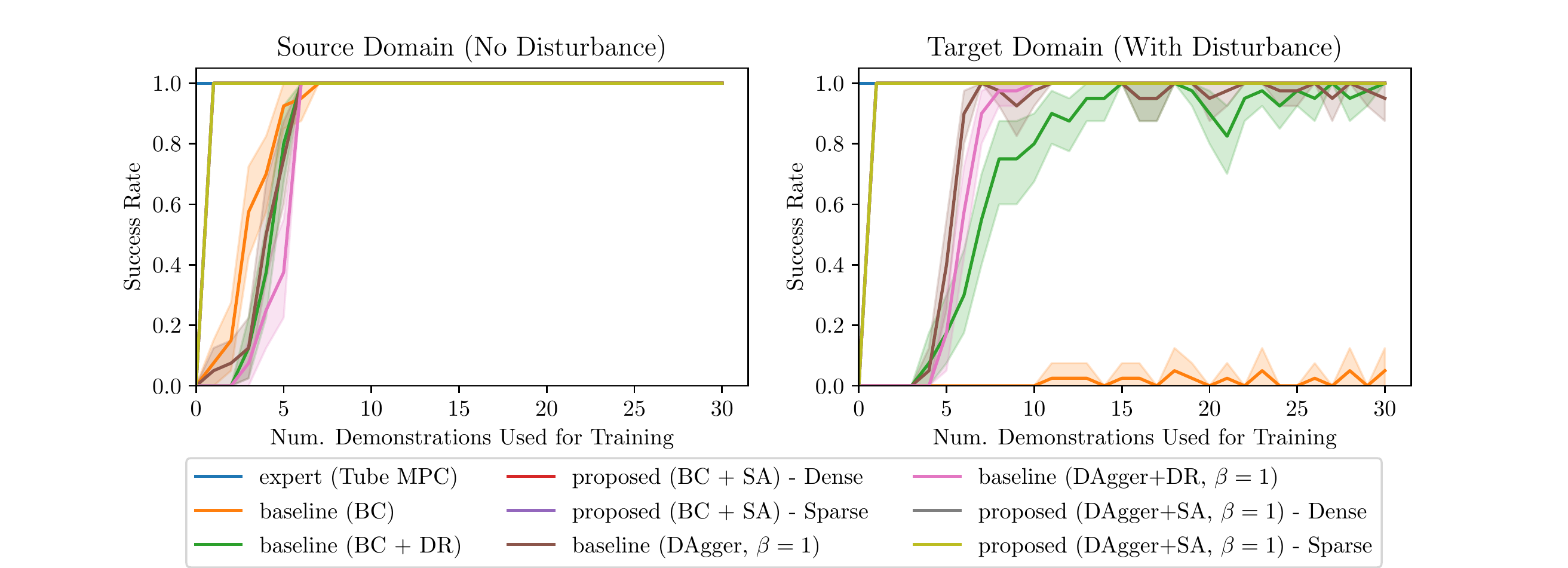}
\end{subfigure}%
    \caption{Robustness (\textit{Success Rate}) in the task of flying along an eight-shaped, $7$s long-trajectory, subject to wind-like disturbances (left) and without (right), starting from different initial states (Task T1). Evaluation is repeated across $8$ random seeds, $20$ times per demonstration per seed. We additionally show the $95\%$ confidence interval.}
    \label{fig:single_trajectory_eval}
\end{figure}

\begin{table}[t]
\newcolumntype{P}[1]{>{\centering\arraybackslash}p{#1}}
    \tiny
    \renewcommand{\tabcolsep}{1pt}
    \centering
    \begin{tabular}{|p{0.9cm}p{0.9cm}||P{0.7cm}|P{0.7cm}|P{0.7cm}|P{0.7cm}|P{0.7cm}|P{0.7cm}|P{0.7cm}|P{0.7cm}|}
    \hline
    \multicolumn{2}{|c||}{Method} &  
    \multicolumn{2}{c|}{Training} & 
    \multicolumn{2}{c|}{\makecell{Robustness\\succ. rate (\%)}} &
    \multicolumn{2}{c|}{\makecell{Performance\\expert gap (\%)}}  &
    \multicolumn{2}{c|}{\makecell{Demonstration\\Efficiency}}\\
    
        Robustif.           & Imitation     & Easy          & Safe          & T1                    &  T2           &  T1           &  T2       & T1        &  T2\\
        \hline
        \hline
        \multirow{2}{*}{-} & BC          & \textbf{Yes}  & \textbf{Yes}      & < 1                  & 100           & 24.15            & 29.47         & -         & 6 \\
                           & DAgger      & \textbf{Yes}  & No                & 98                   & 100           & 15.79    & 1.34       & 7         & 6 \\ 
        \hline
        \multirow{2}{*}{DR}& BC         & No            & \textbf{Yes}      & 95                    & 100           & 10.04             & 1.27        & 14        & 9 \\ 
         & DAgger                       & No            & No                & \textbf{100}          & 100           & \textbf{4.09}    & 1.45        & 10        & 6 \\
        \hline
        \multirow{2}{*}{SA-Dense}      & BC   & \textbf{Yes}  & \textbf{Yes}      & \textbf{100} & 100    & 25.64     & 1.34      & \textbf{1} & \textbf{1} \\
                                                & DAgger        & \textbf{Yes}  & \textbf{Yes$^*$}  & \textbf{100} & 100    & 10.21     & 1.66      & \textbf{1} & \textbf{1} \\
        \hline
        \multirow{2}{*}{\textbf{SA-Sparse}}    & BC     & \textbf{Yes}  & \textbf{Yes}      & \textbf{100} & 100    & 4.23 & \textbf{1.13}    & \textbf{1} & \textbf{1} \\
                                        & \textbf{DAgger} & \textbf{Yes}  & \textbf{Yes$^*$}   & \textbf{100} & 100    & \textbf{3.75} & \textbf{1.07}    & \textbf{1} & \textbf{1} \\
    \hline
    \end{tabular}%
    
    \caption{Comparison of the \ac{IL} methods considered for RT-MPC compression. 
    T1 refers to the trajectory tracking task under wind-like disturbances, T2 under model errors (drag coefficient mismatch). At convergence (iteration 20-30) we evaluate, in the target domain, robustness (success rate) and performance (relative percent error between actions of the expert and of the compressed policy). Demonstration-Efficiency represents the number of demonstrations required to achieve for the first time full success rate. An approach is considered easy if it does not require to apply disturbances/perturbations during training (e.g., in \textit{lab2real} transfer); an approach is considered safe if does not execute actions that may cause state constraints violations (crashes) during training. *Safe in our numerical evaluation, but not guaranteed.}
    \label{tab:comparison}
    \vskip-4ex
\end{table}

\textbf{Tasks description.}
Our objective is to compress an \ac{RTMPC} policy capable to track a $7$s long, eight-shaped trajectory. We evaluate the considered approaches in two different target domains, with wind-like disturbances (T1) or model errors (T2). Disturbances in T1 are sampled adversarially from $\mathbb{W}$ ($\approx$25--30\% of the UAV weight), while model errors in T2 are applied via mismatches in the drag coefficients used between training and testing. 

\textbf{Results.} We start by evaluating the robustness in T1 as a function of the number of demonstrations collected in the source domain.
The results are shown in Fig \ref{fig:single_trajectory_eval}, highlighting that:
\begin{inparaenum}[i)]
\item while all the approaches achieve robustness (full success rate) in the source domain, \ac{SA} achieves full success rate after only a single demonstration, being 5-6 times more sample efficient than the baseline methods;
\item \ac{SA} is able to achieve full robustness in the target domain, while baseline methods do not fully succeed, and converge at much lower rate. 
\end{inparaenum}
These results remark the presence of a distribution shift between the source and target, which is not fully compensated by baselines methods such as \ac{BC}, due to lack of exploration and robustness.
The performance evaluation and additional results are summarized in \cref{tab:comparison}. We highlight that in the target domain, sparse \ac{SA} combined with \ac{DAgger} achieves closest performance to the expert.
Dense \ac{SA} suffers from performance drops, potentially due to the limited capacity of the considered \ac{DNN} or challenges in training introduced by this data augmentation. Because of its effectiveness and greater computational efficiency, we use  sparse \ac{SA} for the rest of the work. 
\cref{tab:comparison} additionally presents the results for task T2. Although this task is less challenging (i.e., all the approaches achieve full robustness), the proposed method (sparse \ac{SA}) achieves highest demonstration efficiency and lowest expert gap, with similar trends as T1. 

\textbf{Computation.} The average latency (on \textsc{i7-8750H} laptop with \textsc{NVIDIA GTX1060 GPU}) for the expert (MATLAB) is $55.3$ms, while for the compressed policy (PyTorch) is $0.25$ms, achieving a two-orders of magnitude improvement. The average latency for the compressed policy on an Nvidia TX2 CPU (PyTorch) is $1.66$ms.%

\begin{figure}[t]
    \centering
    \begin{subfigure}{0.69\columnwidth}
        \centering
        \includegraphics[height=3.0cm, trim={0.5cm 0.05cm 0.6cm 0.2cm},clip]{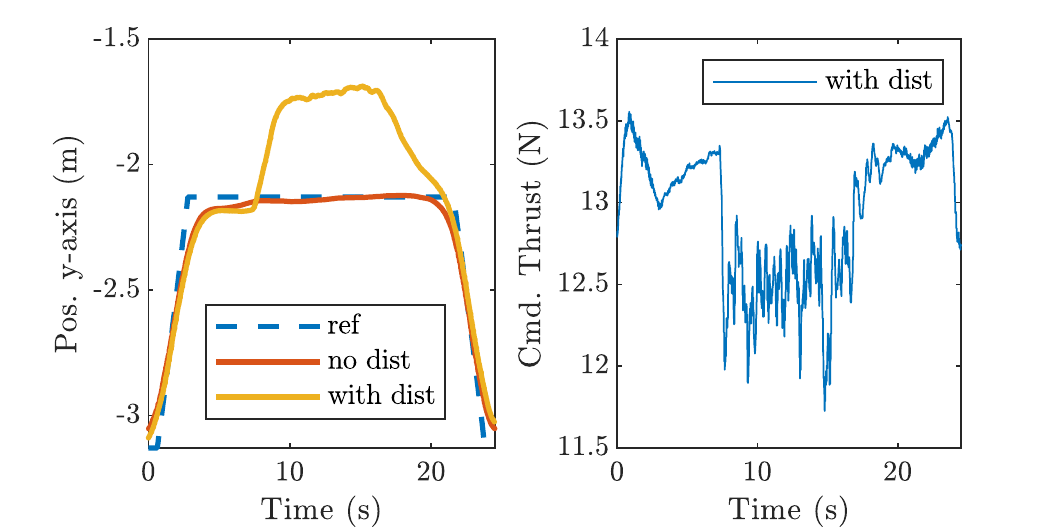}
    \end{subfigure}
    \begin{subfigure}{0.29\columnwidth}
        \centering
        \includegraphics[height=3.0cm, trim={0cm 0 0cm 0},clip]{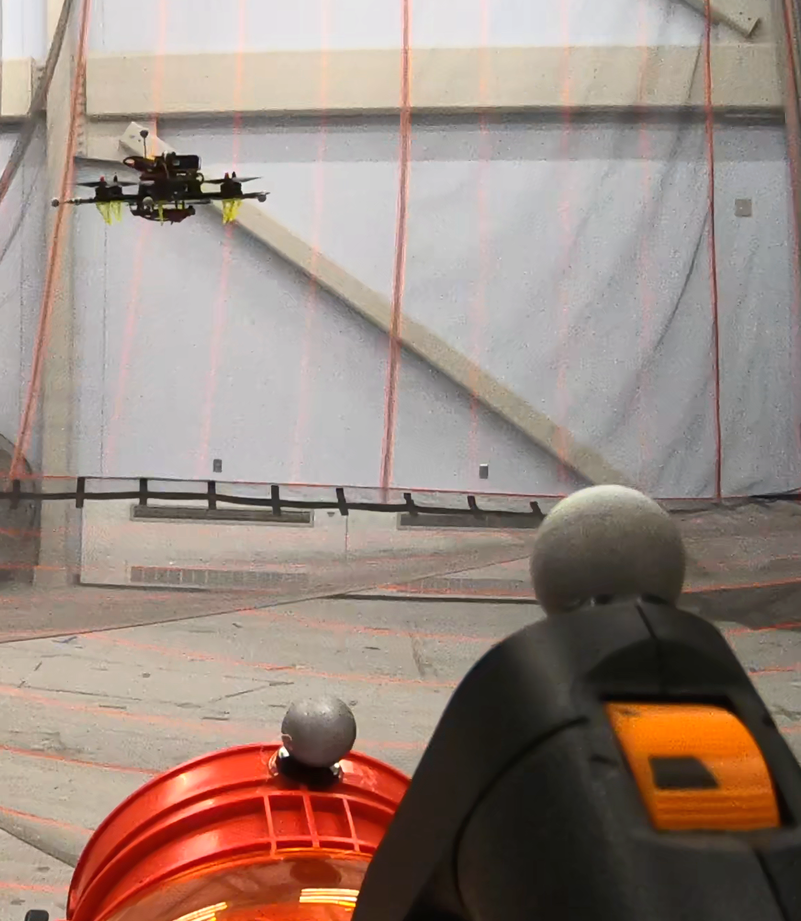}
    \end{subfigure}%
    \caption{Experimental evaluation performed by hovering with and without wind disturbances produced by a leaf blower. The employed compressed RTMPC policy is trained in simulation from a \textit{single} demonstration of the desired trajectory. The wind-like disturbances produce a large position error, but do not destabilize the system. The thrust decreases due to the robot being pushed up by the disturbances. The state estimate (shown in the plot) is provided by onboard \ac{VIO}. The $y$-axis points approximately in the same direction as the wind. 
    }
    \label{fig:experimental_robustness_evaluation_hovering}
\vskip-2ex
\end{figure}
\vskip-2ex

\subsection{Hardware evaluation for tracking a single trajectory}

\begin{figure*}
\captionsetup[sub]{font=footnotesize}
\centering
\begin{subfigure}{0.2\paperwidth}
    \centering
    \includegraphics[height=3.5cm, trim={0cm 0 0cm 0},clip]{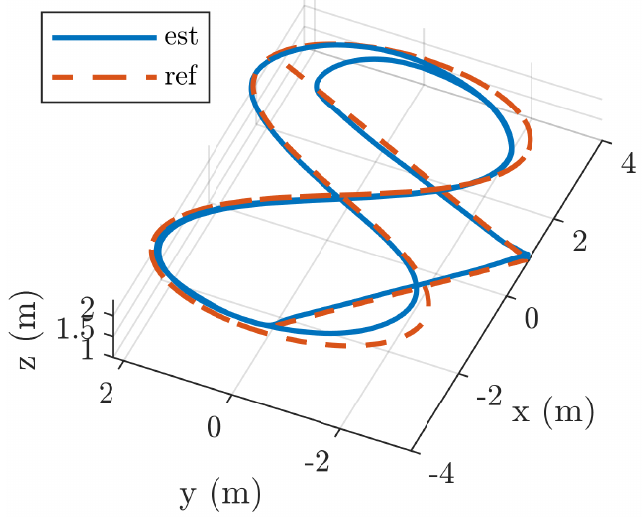}
    \caption{Reference and actual trajectory}
    \label{fig:reference_and_actual_trajectory}
\end{subfigure}%
\hspace{0.1cm}
\begin{subfigure}{0.38\paperwidth}
    \centering
    \includegraphics[height=3.5cm, trim={0.5cm 2cm 1cm 2cm}, clip]{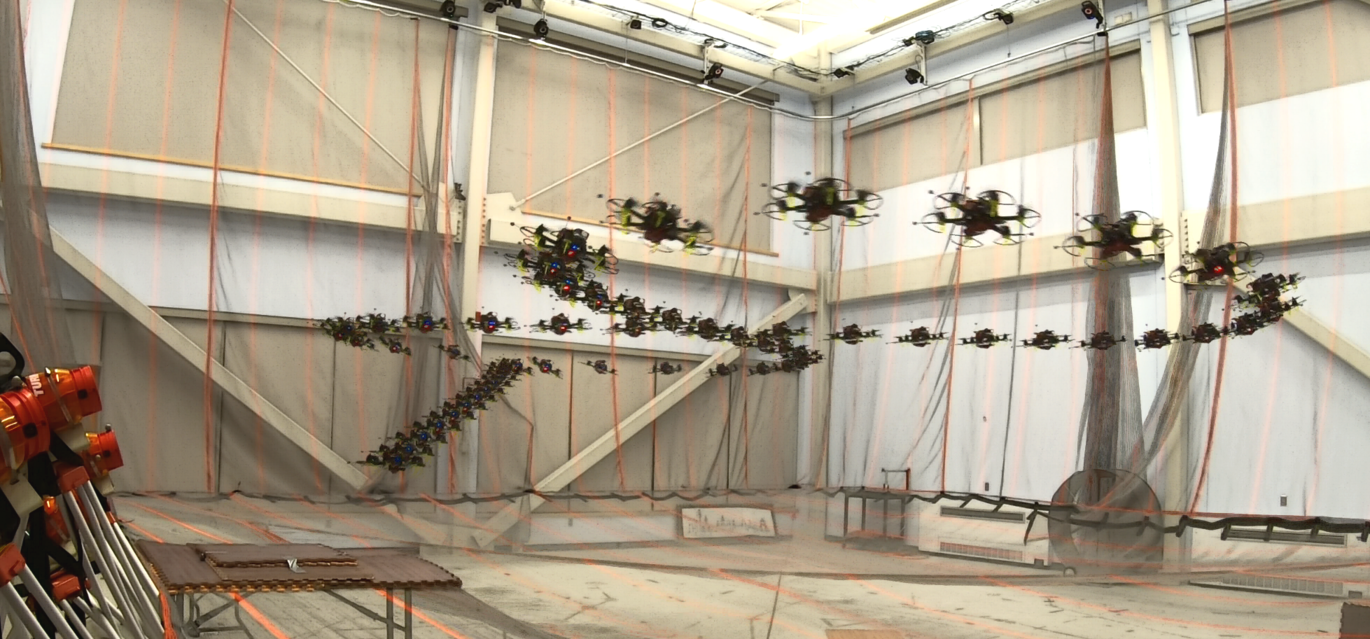}
    \caption{Experimental setup showing the trajectory executed by the robot, and the leaf blowers used to generate disturbances}
    \label{fig:experiment}
\end{subfigure}%
\hspace{0.12cm}
\begin{subfigure}{0.20\paperwidth}
    \centering
    \includegraphics[height=3.5cm, trim={0cm, 0.2cm, 0.5cm, 0.4cm}]{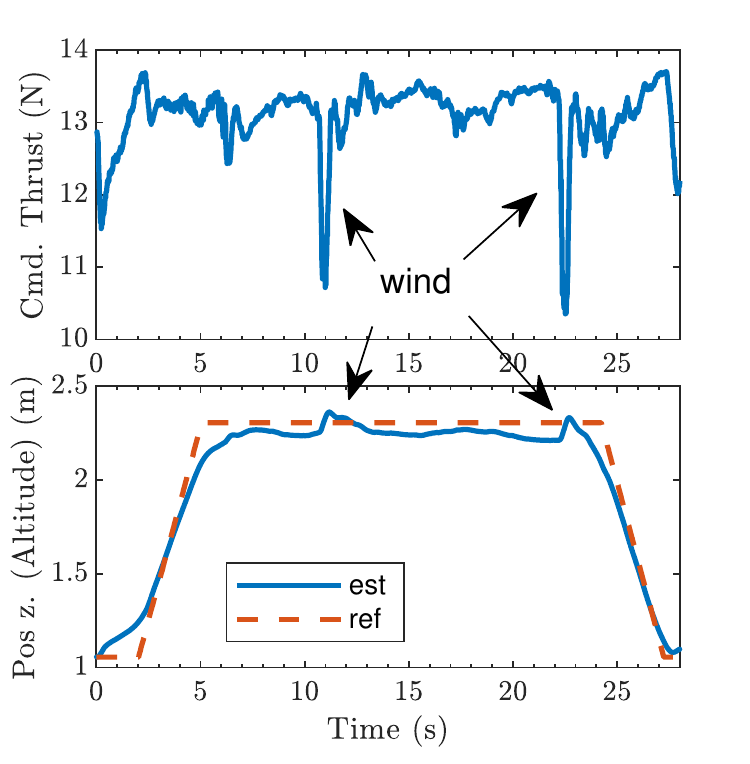}
    \caption{Effects of wind}
    \label{fig:thrust_and_altitude}
\end{subfigure}
    \caption{Experimental evaluation of the proposed approach (\cref{fig:experiment}) under wind-like disturbances. The robot learns to track an eight-shaped reference trajectory (\cref{fig:reference_and_actual_trajectory}) from a \textit{single} demonstration collected in simulation, achieving zero-shot transfer. It is additionally able to withstand disturbances produced by an array of leaf-blowers, unseen during the training phase, and whose effects are clearly visible in the altitude errors (and change in commanded thrust) in \cref{fig:thrust_and_altitude}.}
    \label{fig:experiment_lemniscate}
    \vskip-2ex
\end{figure*}
We validate the demonstration efficiency, robustness and performance of the proposed approach by experimentally testing policies trained after a \textit{single} demonstration collected in simulation using DAgger/BC (which operate identically since we use DAgger with $\beta=1$ for the first demonstration). The data augmentation strategy is based on the sparse \ac{SA}. We use the MIT/ACL open-source snap-stack \cite{acl_snap_stack} for controlling the attitude of the MAV, while the compressed \ac{RTMPC} runs at $100$Hz on the onboard Nvidia TX2 (on its CPU), with the reference trajectory provided at $100$Hz. State estimation is obtained via a motion capture system or onboard \ac{VIO}. The first task considered is to hover under wind disturbances produced by a leaf blower. The results are shown in \cref{fig:experimental_robustness_evaluation_hovering}, and highlight the ability of the system to remain stable despite the large position error caused by the wind. The second task is to track an eight-shaped trajectory, with velocities up to $3$m/s. We evaluate the robustness of the system by applying a wind-like disturbance produced by an array of $3$ leaf blowers (\cref{fig:experiment_lemniscate}). The given position reference and the corresponding trajectory are shown in \cref{fig:reference_and_actual_trajectory}. The effects of the wind disturbances are clearly visible in the altitude errors and changes in commanded thrust in \cref{fig:reference_and_actual_trajectory} (at $t=11$s and $t=23$s). 
These experiments show that the controller can robustly track the desired reference, withstanding challenging perturbations unseen during the training phase. The video submission presents more experiments, including lab2real transfer. 

\begin{figure}
\captionsetup[sub]{font=footnotesize}
\centering
\begin{subfigure}{\columnwidth}
    \centering
    \includegraphics[width=\columnwidth, trim={2cm 2cm 2cm 0},clip]{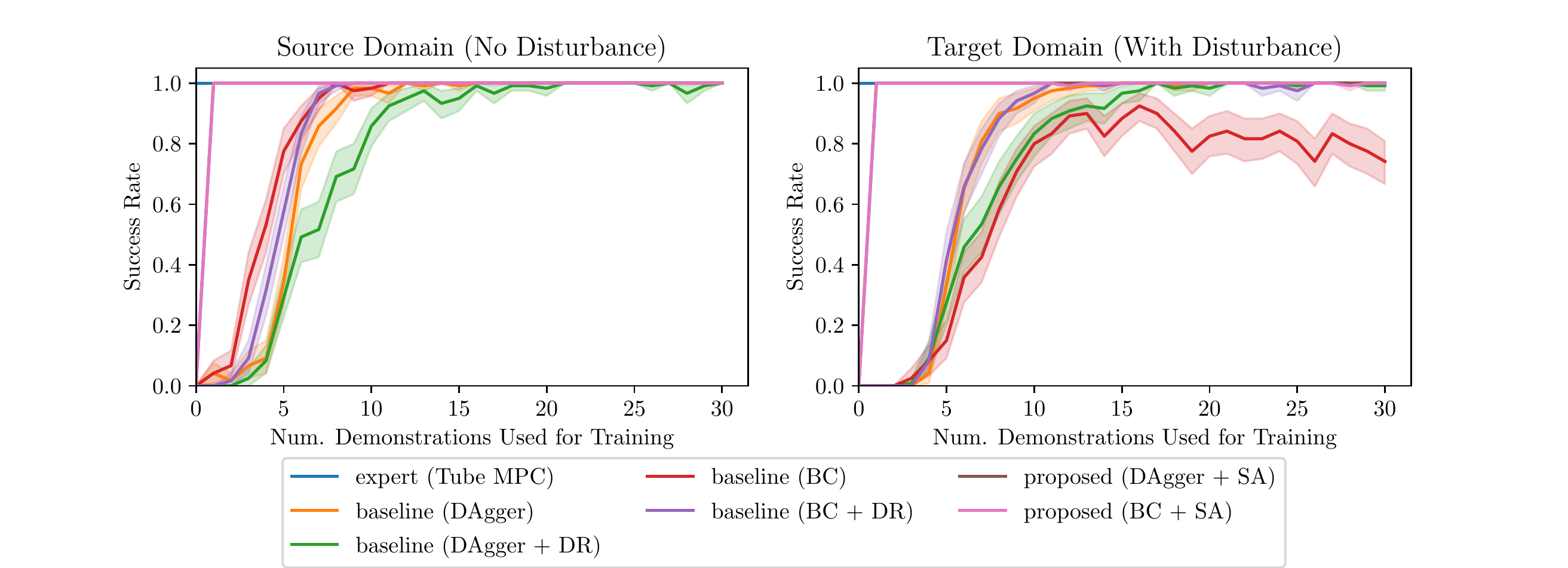}
\end{subfigure}%
\hspace{0.1cm}
\begin{subfigure}{\columnwidth}
    \centering
    \includegraphics[width=\columnwidth, trim={2cm 0 2cm 0},clip]{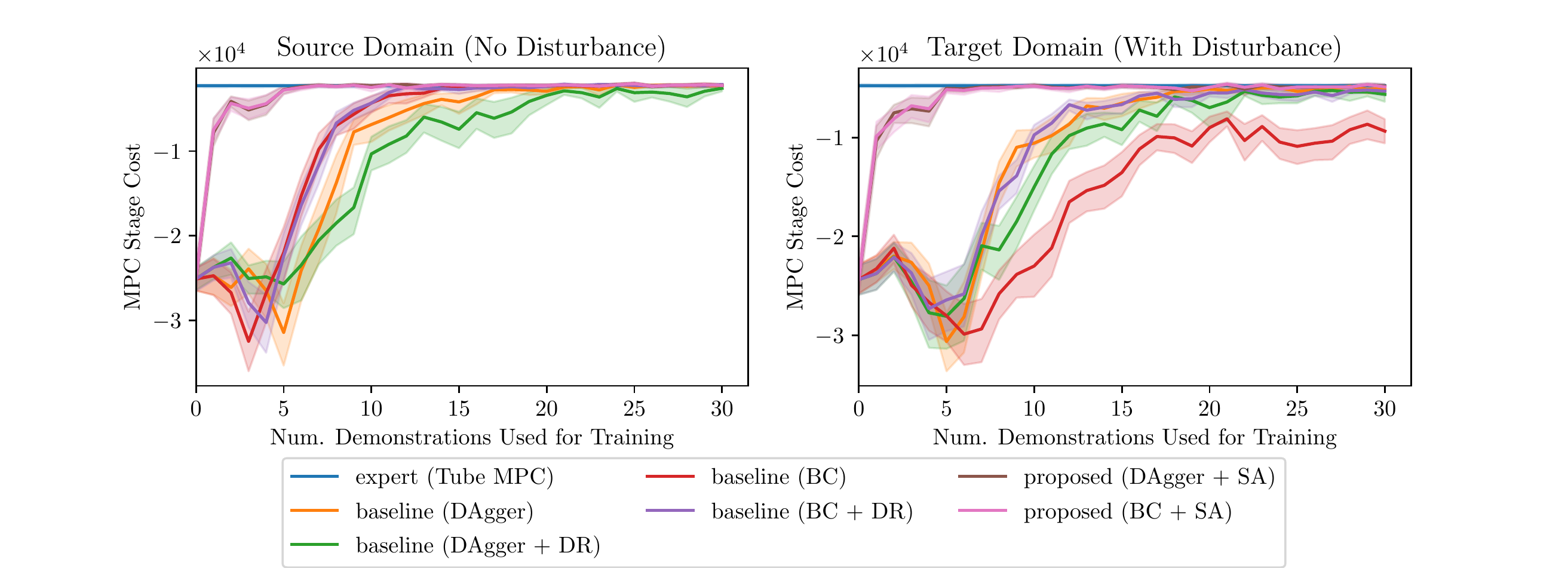}
\end{subfigure}
    \caption{Robustness (\textit{Success Rate}, top row) and performance (\textit{MPC Stage Cost}, bottom row) of the proposed approach (with $95\%$ confidence interval), as a function of the number of demonstrations used for training, for the task of learning to track previously unseen circular, eight-shaped and constant position reference trajectories, sampled from the same training distribution. The left column presents the results in the training domain (no disturbance), while the right column under wind-like perturbations (with disturbance). The proposed RTMPC driven sparse SA strategy learns to track multiple trajectory and generalize to unseen ones requiring less demonstrations. At convergence (from demonstration $20$ to $30$), DAgger+SA achieves the closes performance to the expert ($2.7\%$ \textit{expert gap}), followed by BC+SA ($3.0\%$  \textit{expert gap}). Evaluation performed using $20$ randomly sampled trajectories per demonstration, repeated across 6 random seeds, with prediction horizon of $N=20$ to speed-up demonstration collection, and the DNN input size is adjusted accordingly.}
    \label{fig:learning_multiple_trajectories}
    \vskip-2ex
\end{figure}
\begin{figure}
    \centering
    \includegraphics[width=\columnwidth, trim={1cm, 0.5cm, 1.0cm, 0.5cm}, clip]{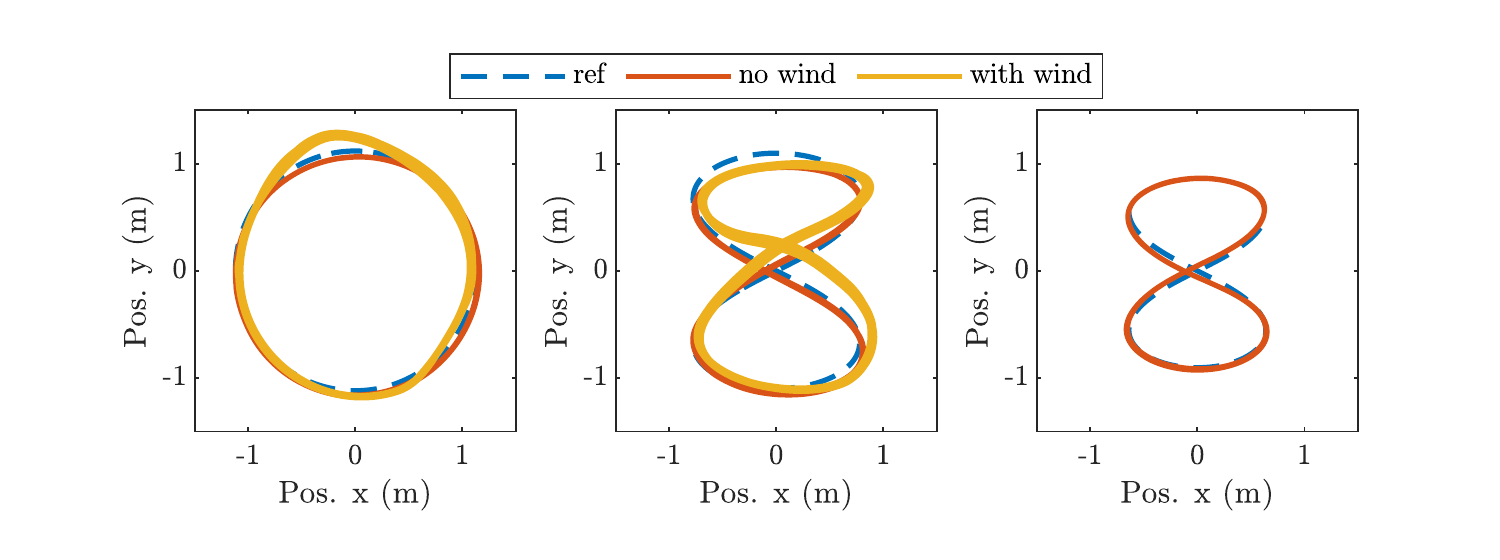}
    \caption{Examples of different, arbitrary chosen trajectories from the training distribution, tested in hardware experiments with and without strong wind-like disturbances produced by leaf blowers. The employed policy is trained with $10$ demonstrations (when other baseline methods have not fully converged yet, see \cref{fig:learning_multiple_trajectories}) using DAgger+SA (sparse). This highlights that sparse SA can learn multiple trajectories in a more sample-efficient way than other IL methods, retaining RTMPC's robustness and performance. Prediction horizon used is $N=20$, and DNN input size is adjusted accordingly.}
    \label{fig:learn_multiple_trajectories_experiment}
    \vskip-2ex
\end{figure}
\subsection{Numerical and hardware evaluation for learning and generalizing to multiple trajectories}
We evaluate the ability of the proposed approach to track multiple trajectories while generalizing to unseen ones. To do so, we define a training distribution of reference trajectories (circle, position step, eight-shape) and a distribution for these trajectory parameters (radius, velocity, position). 
During training, we sample at random a desired, $7$s long ($70$ steps) reference with randomly sampled parameters, generating a demonstration and updating the proposed policy, while testing on a set of $20$, $7$s long trajectories randomly sampled from the defined distributions. We monitor the robustness and performance of the different methods, with force disturbances applied in the target domain. The results of the numerical evaluation, shown in \cref{fig:learning_multiple_trajectories}, confirm that sparse SA
\begin{inparaenum}[i)]
    \item achieves robustness and performance comparable to the one of the expert in a sample efficient way, requiring less than half the demonstrations than baseline approaches; 
    \item simultaneously learns to generalize to multiple trajectories randomly sampled from the training distribution.
\end{inparaenum}
The hardware evaluation, performed with DAgger augmented via sparse SA, is shown in \cref{fig:learn_multiple_trajectories_experiment}. It confirms that the proposed approach is experimentally capable of tracking multiple trajectories under real-world disturbances/model errors.

\section{Conclusion and Future Work}

This work has presented a demonstration-efficient strategy to compress a \ac{MPC} in a computationally efficient representation, based on a \ac{DNN}, via \ac{IL}. 
We showed that greater sample efficiency and robustness than existing \ac{IL} methods (\ac{DAgger}, \ac{BC} and their combination with \ac{DR}) can be achieved by designing a Robust Tube variant of the given MPC, using properties of the tube to guide a sparse data augmentation strategy. Experimental results -- showing trajectory tracking control for a multirotor after a \textit{single} demonstration under wind-like disturbances -- confirmed our numerical findings. Future work will focus on designing an adaptation strategy.

\section*{ACKNOWLEDGMENT}
This work was funded by the Air Force Office of Scientific Research MURI FA9550-19-1-0386. 

\balance
\bibliographystyle{IEEEtran}
\bibliography{main}

\begin{acronym}
\acro{OC}{Optimal Control}
\acro{LQR}{Linear Quadratic Regulator}
\acro{MAV}{Micro Aerial Vehicle}
\acro{GPS}{Guided Policy Search}
\acro{UAV}{Unmanned Aerial Vehicle}
\acro{MPC}{Model Predictive Control}
\acro{RTMPC}{Robust Tube MPC}
\acro{DNN}{deep neural network}
\acro{BC}{Behavior Cloning}
\acro{DR}{Domain Randomization}
\acro{SA}{Sampling Augmentation}
\acro{IL}{Imitation Learning}
\acro{DAgger}{Dataset-Aggregation}
\acro{MDP}{Markov Decision Process}
\acro{VIO}{Visual-Inertial Odometry}
\end{acronym}

\end{document}